\documentclass[a4paper]{article}
\usepackage{cite}
\usepackage{url} 
\usepackage{hyperref}       
\usepackage[utf8]{inputenc}
\usepackage{amsmath}
\usepackage{amsfonts}
\usepackage{amssymb}
\usepackage{graphicx}
\usepackage{xcolor}
\usepackage{enumitem}
\usepackage{lineno}

\definecolor{darkgreen}{rgb}{0.1,.8,0.1}
\definecolor{darkred}{rgb}{0.8,.1,0.1}
\definecolor{darkblue}{rgb}{0.1,.1,0.8}

\newcommand{\costmfg}{\mathcal{C}}
\newcommand{\costmfgx}{\mathcal{C}^x}
\newcommand{\costmfgy}{\mathcal{C}^y}
\newcommand{\costmfgf}[2]{\mathcal{C}(#1, #2)}
\newcommand{\costmfgfx}[2]{\mathcal{C}^x(#1, #2)}
\newcommand{\costmfgfy}[2]{\mathcal{C}^y(#1, #2)}
\newcommand{\division}[2]{\mathcal{S}(#1, #2)}
\newcommand{\espacex}{\mathcal{X}}
\newcommand{\espacey}{\mathcal{Y}}

\newcommand{\varx}{\Upsilon}
\newcommand{\varxx}{\Upsilon^x}
\newcommand{\varxy}{\Upsilon^y}

\newcommand{\R}{{\mathbb R}}

\newcommand{\probasetp}[2]{{\mathcal P}_#2(#1)}

\newcommand{\E}{{\mathbb E}}

\def\beq{\begin{equation}}
\def\eeq{\end{equation}}
\def\beqn{\begin{eqnarray}}
\def\eeqn{\end{eqnarray}}
\newcommand{\abs}[1]{\left\lvert#1\right\rvert}

\newcommand{\bartau}{\bar{\tau}}
\newcommand{\barx}{\bar{x}}
\newcommand{\bary}{\bar{y}}
\newcommand{\minimizer}{\mathcal{M}}

\newcommand{\bproof}{{\bf Proof }}
\newcommand{\eproof}{{\hfill $\blacksquare$}}
\newtheorem{remark}{Remark}
\newtheorem{theorem}{Theorem}

\newtheorem{lemma}{Lemma}

\newtheorem{definition}{Definition}

\title{Convergence dynamics of Generative Adversarial Networks: the dual metric flows}

\author{Gabriel Turinici\footnote{ORCID: 0000-0003-2713-006X, \text{Gabriel.Turinici@dauphine.fr}, \text{http://www.turinici.com}} \\
CEREMADE, Universit\'e Paris Dauphine - PSL,	
 Paris, France}
\date{Jan 10,2021}
\begin{document}
\maketitle
\begin{abstract}
Fitting neural networks often resorts to stochastic (or similar) gradient descent which is a noise-tolerant (and efficient)
resolution of a gradient descent dynamics. It outputs a sequence of networks parameters, which sequence evolves during the 
training steps. The gradient descent is the limit, when the learning rate is small and the batch size is infinite,
of this set of increasingly optimal network parameters obtained during training.
In this contribution, we investigate instead the convergence in the Generative Adversarial Networks used in machine learning. We study the limit of small learning rate, and show that, similar to single network training, the GAN learning dynamics tend, for vanishing learning rate
to some limit dynamics.
This leads us to consider evolution equations in metric spaces (which is the natural framework for  evolving
probability laws)
that we call dual flows. 
We give formal definitions of solutions and prove the convergence.
The theory is then applied to specific instances of GANs and we discuss how this insight helps understand and mitigate the mode collapse.
	
\end{abstract}


\section{Introduction}

Deep generative models are of high interest and used in many applications of deep learning. Among them, the GANs have been one of the most efficient in terms of practical results. 
The GANs and their convergence are the object of a huge quantity of research papers ($4'861$ arxiv results mid-October 2020 for "generative adversarial network", 
$26'110$ Google Scholar results). 
Nevertheless, only very few works concern the behavior of solutions in the general framework of metric spaces or the meaning to be given to the learning trajectory in the limit 
of a small learning rate.
On the other hand,
the GANs are known to exhibit unstable convergence behavior (see~\cite{kodali_convergence_2017})
and several procedures have been proposed to cure this drawback, among which~\cite{arjovsky_wasserstein_2017,gulrajani_improved_2017,deshpande_max-sliced_2019,wu_sliced_2019,kolouri_sliced-wasserstein_2018}. In order to contribute to a 
fundamental understanding of the objects involved, we give in this work a rigorous definition of the concept of solution of 
the evolution equation associated to a GAN that we call a {\it dual metric flow}.
We identify the hypothesis that guarantee that the discrete solutions 
converge, when the learning rate $\tau$ tends to  $0$, to a dual metric flow and apply this insight to understand and mitigate the mode collapse 
phenomena. 
Finally we give examples that show that the dual flows correspond indeed to procedures 
used in GAN practice.

\subsection{Motivation: W-GANs}

\begin{figure}[htb!]
\includegraphics[width=0.45\textwidth]{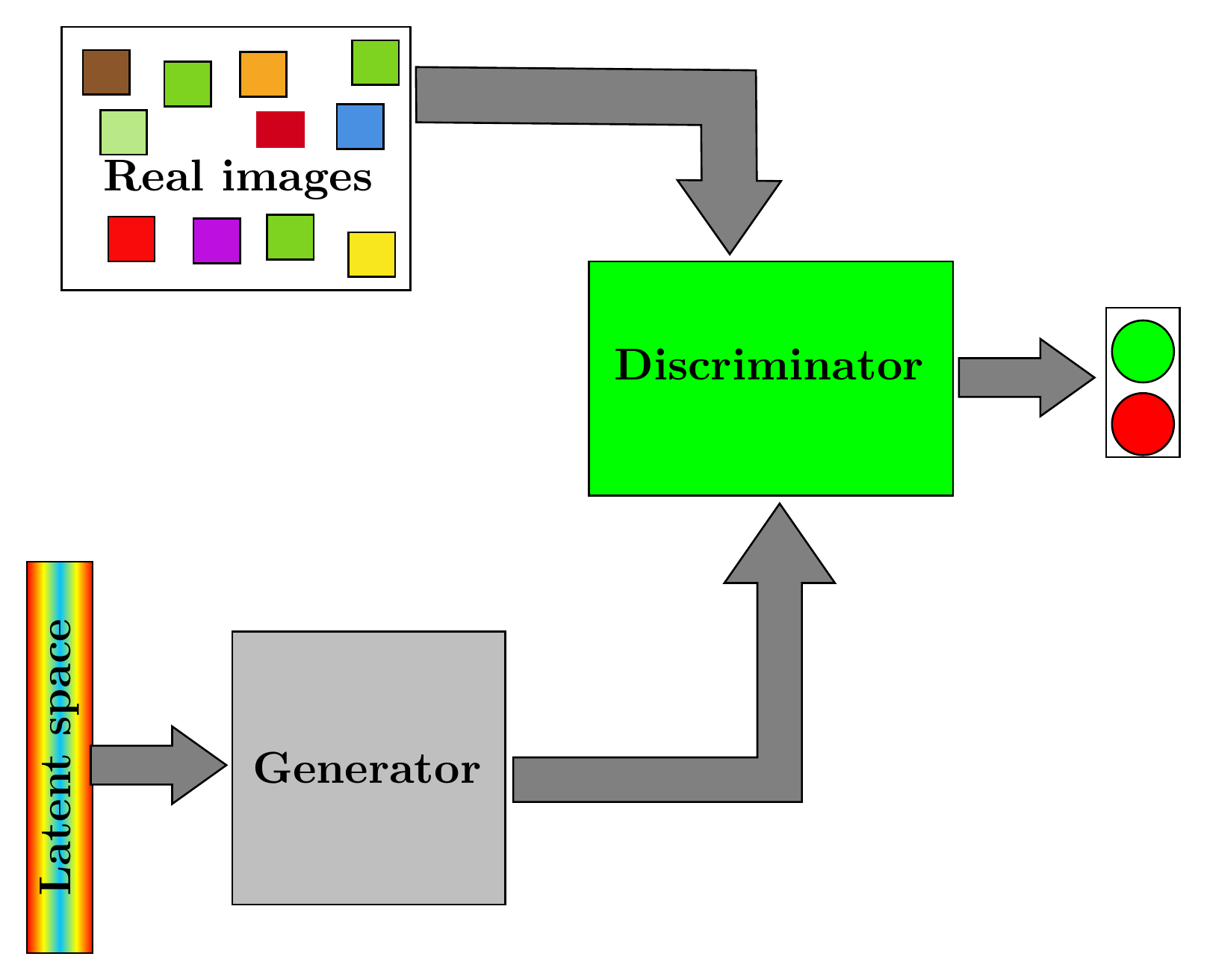}
\includegraphics[width=0.45\textwidth]{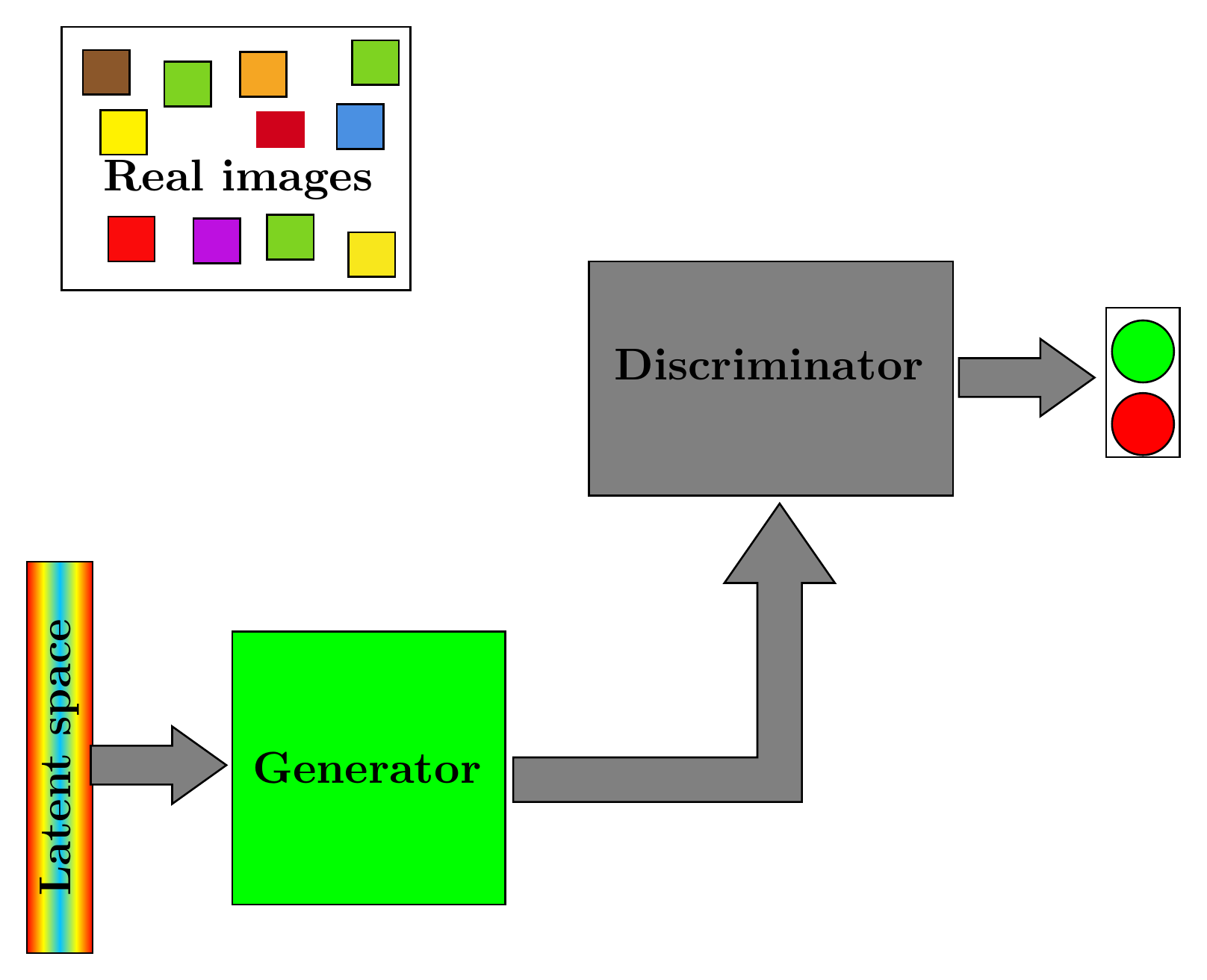}
\caption{Illustration of the dynamics of a GAN. Color code: in green the part that is active (under training)
and in grey the part that is fixed. {\bf Left:} the discriminator is active. {\bf Right:} the generator is active. This
is repeated till convergence.}
\label{fig:defgan}
\end{figure}

The goal of (deep) generative models such as the GANs is to generate new data from some (unknown) distribution given a list of samples drawn from that distribution.
To simplify the presentation, we suppose that the distribution to be learned is a set of images. 
Two objects are important in a GAN: the {\it Generator} and the {\it Discriminator}; both are deep networks with fixed, 
but rich enough, architecture capable of representing a very large class of transformations. For instance, in a Wasserstein-GAN
(see~\cite{arjovsky_wasserstein_2017}), the training has the following form (see figure~\ref{fig:defgan} for an illustration): after initializing (randomly) both the Generator and the Discriminator, the Discriminator is trained first. It takes as input images generated by the Generator (with label "fake") and images from the real database with
label "true". It is trained for (one or possibly several) steps in order to achieve good discrimination efficiency between the "fake" and "real" labels. In the next step the discriminator is kept constant and the generator is 
trained in order to create images which, when run through the (fixed) discriminator obtain as much labels "real" as possible.
Then the procedure is repeated till convergence.

\begin{figure}[htb!]
	\includegraphics[width=0.49\textwidth]{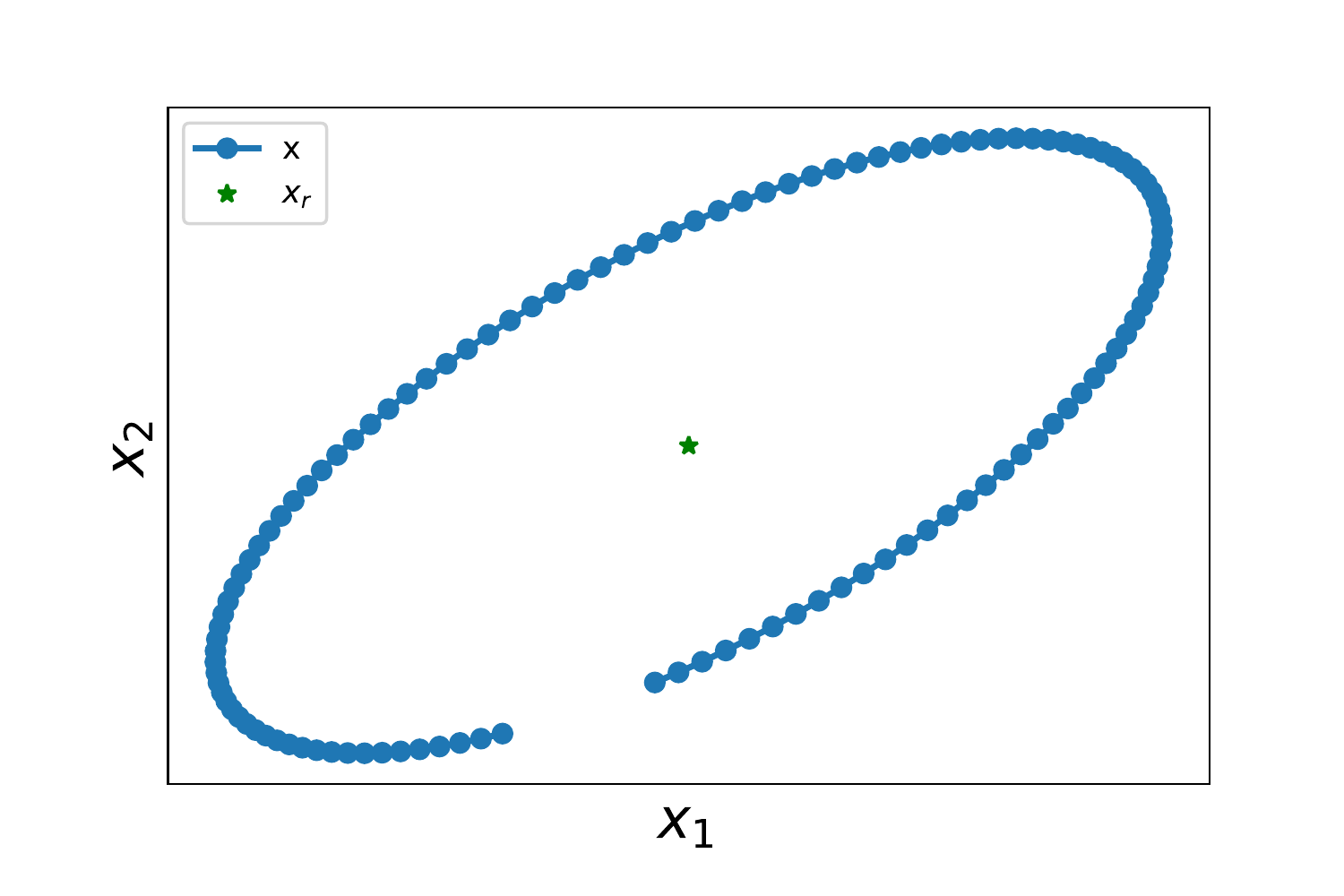}
	\caption{Illustration of the oscillations dynamics of a GAN, see equation \eqref{eq:gansimple}.}
	\label{fig:ganosc}
\end{figure}

A very natural question is whether such a procedure can converge to a satisfactory solution i.e., if the Generator  samples from the right distribution and the Discriminator is able to tell with high precision the quality of any sample.
The answer is not always yes, as illustrated by the following simple situation: consider a target distribution which is a Dirac mass centered in some constant $x_{r}$. The generator is described by a vector of two real parameters 
$x \in \R^2$ and the discriminator $y$ has the same format. 
In the framework of integral probability metrics (see~\cite{sriperumbudur_hilbert_2010}) used in GANs, this simple situation has the following transcription: if the current parameter of the generator is $x_n$ then 
the next parameter $y_{n+1}$ of the discriminator will be updated to maximize the distance from $x_n$ to $x_r$, i.e.
 $y_{n+1} = y_n + \tau (x_n - x_r)$, where $\tau$ is the learning rate. The generator itself
 will take into account this new discriminator and will move towards the unknown value $x_r$ by taking a step:
 $x_{n+1} = x_n - \tau y_{n+1}$, where $\tau$ is the learning rate. These equations can be
 also written as
 \begin{eqnarray}
 & \ & 
\frac{ y_{n+1}- y_n}{\tau}  = x_n - x_r \nonumber\\
 & \ & 
\frac{ x_{n+1}- x_n}{\tau}  = y_{n+1}.  \label{eq:gandiscrete0}
 \end{eqnarray}
 When $\tau \to 0$ the limit evolution will be 
 \begin{eqnarray}
 & \ & 
 y'(t) = x(t) - x_r \nonumber\\
 & \ & 
x'(t)  = y(t). \label{eq:gansimple}
 \end{eqnarray}
However, except for very special initial conditions, the system~\eqref{eq:gansimple} 
does not have the property that $x(t)\to x_r$ because $x(t)$
will have a periodic evolution around $x_r$, see figure~\ref{fig:ganosc}.

Prompted by this example we aim to analyze in this paper the intrinsic constraints coming from the GAN training in the form of an alternative adversary evolution. More precisely, given the discrete Generator / Discriminator dynamics (similar to equation \eqref{eq:gandiscrete0}) we want to write the equivalent limit equation \eqref{eq:gansimple}.

To do so, we suppose that the architectures of the Generator and Discriminator networks are rich enough so that the Generator can reach with satisfactory precision any target distribution $\mu_r$ and the discriminator can realize any mapping that separates an arbitrary pair of distinct distributions in the integral probability metrics sense, i.e. for any distinct distributions the Discriminator can propose a mapping whose averages under the two distributions are different. Of course this is an ideal setting but we take this view in order to better distinguish the effects due to network architectures from those intrinsically included in the GAN convergence protocol.

Let us now introduce the mathematical objects involved in the GAN training. The Generator is a mapping from 
a given distribution (e.g. the multi-dimensional Gaussian distribution) on some base space (called Latent space) to the 
space of objects of interest, denoted $\Omega$; for instance in figure~\ref{fig:defgan} $\Omega$ are images).
Thus in general the Generator can construct probability distributions on $\Omega$; denote by $\probasetp{\Omega}{1}$
the set of all probability laws on $\Omega$ with finite first order moment. This set can be given the structure of a metric space by introducing a distance; many distances exists and have been used but a popular choice, used by W-GANs is the $1$-Wasserstein distance (see~\cite{ambrosio_gradient_2008} for a definition) denoted $d_{W,1}$. Thus formally we work in the metric space $(\probasetp{\Omega}{1},d_{W,1})$. 

On the other hand the Discriminator constructs a Lipschitz mapping to label the generated samples as fake or real. Mathematically the object is a 
Lipschitz function $\ell :\Omega \to \R$. The Discriminator works
best when the average $\E_\mu(\ell)$ is as different as possible from the average  $\E_{\mu_r}(\ell)$. Again, $\ell$ lives in a metric space $\espacey$ (as opposed to a Hilbert space) with the distance being the Lipschitz distance.

The GAN update of the Generator  will move $\mu_n$ ($n$ indexes now the learning steps) to some $\mu_{n+1}$ along the gradient of the mapping 
$\mu \mapsto \E_\mu(\ell)-\E_{\mu_r}(\ell)$ (to be minimized, here $\ell$ being the current Discriminator state). 
In practice the update is performed with a stochastic descent algorithm (that we take in this work to be the SGD). We will not inquire about the stochastic oscillations
but only consider the average state.
In full rigor the update rule is difficult to write in the formal, metric space setting: for instance, for the Generator update, the intuitive formula would be 
$ \mu_{n+1} = \mu_n - \tau \nabla_\mu \left[\E_{\mu_n}(\ell_{n+1})-\E_{\mu_r}(\ell_{n+1})\right]$. This formulation has several problems: first the space of 
probability laws is not a vector space thus the addition and substractions operator are not well defined. Secondly, the differential structure on the same space
is not straightforward to manipulate, i.e., $\nabla_\mu$ is not easy to work with. For these reasons, we will replace the real dynamics with another, close, one
and we model the update by the requirement that $\mu_{n+1}$ is the minimizer of

\begin{equation}
\mu \mapsto \frac{d_{\espacex}(\mu_n,\mu)^2}{2 \tau} +  \left[\E_\mu(\ell)-\E_{\mu_r}(\ell)\right].
\label{eq:jko1}
\end{equation}
The intuitive justification for this formula (very classic in metric space evolution equations~\cite{gigli}) is the
following: in a Hilbert space, $d_{\espacex}(\mu_n,\mu)^2 = \| \mu_n - \mu \|^2$ and taking a general functional  
$F(\mu)$ 
in~\eqref{eq:jko1}
(here $F(\mu)=\E_\mu(\ell)-\E_{\mu_r}(\ell)$) 
the critical point equations derived from \eqref{eq:jko1} can be written as  
$ \mu_{n+1} = \mu_n - \tau \nabla F(\mu_{n+1})$
which is an implicit gradient descent of step $\tau$ starting from $\mu_n$. The advantage of the implicit
formulation is that $\mu_{n+1}$ defined through \eqref{eq:jko1} can be written for any abstract objects
in a metric space (only $d_{\espacex}(\mu_n,\mu)^2$ and $F(\cdot)$ are required).
Same model is used for the Discriminator update steps; all this leads us to consider the main discrete equation that 
will model the GAN training dynamics~\footnote{see Lemma \ref{lemma:explicit} in appendix for information on the relationship between explicit and implicit numerical schemes.}

\begin{eqnarray}
& \ & 
\!\!\!\!\!\!\!\!\!\!\!\!\!\!\!\!\!\!\!\!\!\!\!\!
\ell_{n+1} = \textrm{argmin}_{\ell \in \espacey} \frac{d(\ell,\ell_n)^2}{2 \tau } - 
\left[\E_{\mu_n}(\ell)-\E_{\mu_r}(\ell)\right] \\
& \ & 
\!\!\!\!\!\!\!\!\!\!\!\!\!\!\!\!\!\!\!\!\!\!\!\!
\mu_{n+1} = \textrm{argmin}_{\mu \in \espacex} \frac{d(\mu,\mu_n)^2}{2 \tau } + 
\left[\E_\mu(\ell_{n+1})-\E_{\mu_r}(\ell_{n+1})\right],
\end{eqnarray}
with $\tau >0$ the learning rate and $\espacex, \espacey$ two metric spaces.
The goal of this paper is to clarify whether, when $\tau \to 0$, the discrete curves 
$(\mu_n)_{n\ge 0}$,
$(\ell_n)_{n\ge 0}$,
converge to some limit continuous curves; we also want to give a formal definition of the dynamics satisfied by the limit curves.

The GAN protocol described above is not the only one used in the literature. Many other procedures aim to improve convergence,
generation quality, computing speed, etc. 
For instance in \cite{gulrajani_improved_2017} the authors add a penalty on the gradient in order to make convergence better and the new
type of GAN is called WGAN-GP. 
A full zoology of GAN versions appeared: 
the Deep Convolutional Generative Adversarial Network (DCGAN, \cite{radford_unsupervised_2016}) use deep convolution networks 
which are better adapted to images, the Coupled GANs proposal (coGAN \cite{coupled_gan}) uses two generators and two discriminators in order to improve convergence 
and generation properties,
Progressive Growing Generative Adversarial Network (Progressive GAN \cite{karras_progressive_2018}) 
improve the generation quality by gradually increasing image resolution (size), 
Style-Based Generative Adversarial Network (StyleGAN, \cite{stylegan}) better control the Latent space distribution and
are able to generate content with given characteristics; other contributions include 
 Cycle-Consistent Generative Adversarial Network (CycleGAN, \cite{zhu_unpaired_2017}),
Big Generative Adversarial Network (BigGAN, \cite{brock_large_2019}), Pix2Pix \cite{pix2pixGAN}, 
and the research is still advancing.

 In~\cite{wasswassgan} the authors use
even more abstract objects which lead to a loss functional called
"Wasserstein of Wasserstein loss"; that is, the most basic objects that are here the images
 (with our notations elements $ \omega \in \Omega$) are not given 
the usual Euclidian distance, but instead are immersed in a metric space and the Wasserstein distance
$d_{W,1}(\omega_1,\omega_2)$
 is used to measure the 
distance (dissimilarity) 
between two (elementary) images $\omega_1, \omega_2 \in \Omega$. The Wasserstein distance 
$d_{W,1}(\mu_1,\mu_2)$
is 
then used a second time in order to discriminate between probability laws 
 $ \mu_1, \mu_2 \in \probasetp{\Omega}{1}$.

Similar abstraction for the ground distance are to be found in \cite{adler_banach_2018} that present Banach space GANs.

All these examples enforce even more the need for an abstract formulation of GAN convergence, that we give below. 
Finally, see also~\cite{iwaki_implicit_2019} for implicit procedures relevant to GANs.

\subsection{Mathematical setting}

We   consider $\espacex$, $\espacey$  two (Polish geodesic) metric spaces (see \cite{metricbook} for an introduction to metric spaces) and 
 $\costmfgf{\cdot}{\cdot}:\espacex\times \espacey \to \R \times \R$ a functional (that will stand for the loss functional). Note that $\costmfgf{\cdot}{\cdot}$
 is vector valued, we will denote by $\costmfgfx{\cdot}{\cdot}$ and $\costmfgfy{\cdot}{\cdot}$ its components.
Note that in a GAN  we will have 
\begin{eqnarray}
& \ & 
\costmfgf{\ell}{\mu}
= (\costmfgfx{\ell}{\mu},\costmfgfy{\ell}{\mu})
\nonumber \\ & \ & 
=
(-\E_{\mu}(\ell)+\E_{\mu_r}(\ell),\E_{\mu}(\ell)-\E_{\mu_r}(\ell)).
\label{eq:ganfunctions}
\end{eqnarray}

We investigate the equation:
\beq
\label{eq:eqflow0}
\partial_t 
\begin{pmatrix}
	x_t \\ y_t
\end{pmatrix} + 
\begin{pmatrix}
\nabla_x \costmfgfx{x_t}{y_t} \\ 
\nabla_y \costmfgfy{x_t}{y_t}
\end{pmatrix} 
=0, \ 
\begin{pmatrix}
	x_0 \\ y_0
\end{pmatrix} 
= \begin{pmatrix}
	\barx \\ \bary
\end{pmatrix}.
\eeq

Such an equation will be called a {\it dual flow}. The discrete version is defined by the recurrence:	
\beq \label{eq:bestreplydef}
x_0^\tau=\barx, \ 
x_{k+1}^\tau \in arg min_{x \in \espacex} \frac{d(x,x_k^\tau)^2}{2 \tau} + \costmfgfx{x}{y_k^\tau}, \ \ k \ge 0.
\eeq

\beq \label{eq:bestreplydefy}
y_0^\tau=\bary, \ 
y_{k+1}^\tau \in arg min_{y \in \espacey} \frac{d(y,y_k^\tau)^2}{2 \tau} + \costmfgfy{x_{k+1}^\tau}{y}, \ \ k \ge 0.
\eeq

These numerical schemes are a distant cousin of some other evolution evolution on metric spaces, namely the 
evolution flows, see \cite{Blanchet2014,turinici_metric_2017}).

From the theoretical point of view, these results are not available with previous techniques from \cite{Ambrosio2013,Ferreira,evakopfer1,Ferreira,mielke2008,mielke2012,mielke2013}).

\section{Theoretical results}

\subsection{Motivation and literature review}

Note that when $\costmfgx$ is independent of the second argument, i.e., 
\beq \label{eq:gradflow}
\costmfgfx{x}{y}= E(x),
\eeq
 the relation  \eqref{eq:bestreplydef} becomes the celebrated 
  implicit Euler-type scheme of {Jordan, Kinderlehrer and Otto}~\cite{jko} for the definition of gradient flows in metric spaces
 \beq \label{eq:gradflowclassic}
 \partial_t y_t + \nabla E(y_t)=0, \ y_0=\bary,
 \eeq
  and received 
  considerable attention (see~\cite{Villani2009,filippo,gigli} for instance). 
  However, the situation when $E$ has dependence on other variables has not been treated to the same extent and the related contributions 
  involve gradient flows of time dependent functionals $\mathcal{E}(t,u)$ with a known dependence on time (see \cite{evakopfer1,Ferreira,mielke2008,mielke2012,mielke2013}). 
   Of course, formally one can  set 
$\mathcal{E}(t,u)= \costmfgf{u}{y_t}$,
and hope to analyze the $(x_t,y_t)$ dynamics in this way. This is not possible 
for technical reasons (see for instance the  discussion in ~\cite{turinici_metric_2017}); in particular doing so supposes the knowledge of the
dynamics $y_t$ (which is not available) and moreover the dynamics may not be differentiable with respect to time (but remains absolute continuous).

\subsection{Basic reminders}

The absence of a vector operations in a metric space does no allow to develop fully a differential calculus and requires adaptation of notions of derivative.  
 Accordingly the definition of  evolution equations have to use alternative properties. 
 
 We recall below the main ideas of such an alternative formulation 
 (see ~\cite{Ambrosio2013}) for the particular case \eqref{eq:gradflow}-\eqref{eq:gradflowclassic}; suppose for a moment that $\espacex$ is an Euclidian space and
 $E$ a smooth ($C^1$ or above) function; then:
 
 \beqn
 & \ & 
 \frac{d}{dt} E(x_t)= \langle \nabla E(x_t) ,  x_t' \rangle 
 \ge - \abs{ \nabla E(x_t)} \cdot\abs{x_t'}
 \nonumber\\ & \ &  
 \ge
 -\frac{1}{2} \abs{x_t'}^2 -\frac{1}{2} \abs{\nabla E}^2(x_t),
 \nonumber 
 \eeqn
 or equivalently, 
 \beq
 \frac{d}{dt} E(x_t) +\frac{1}{2} \abs{x_t'}^2 +\frac{1}{2} \abs{\nabla E}^2(x_t) \ge 0 \ \forall t,
 \eeq
 with equality only if $x$ is solution of \eqref{eq:gradflowclassic}.
 Therefore asking that  
 \beq \label{eq:equivalencedef01}
 \frac{d}{dt} E(x_t) +\frac{1}{2} \abs{x_t'}^2 +\frac{1}{2} \abs{\nabla E}^2(x_t) \le 0 \ \forall t,
 \eeq
 is an equivalent characterization of \eqref{eq:gradflowclassic} (more precisely called the EDI formulation). Its integral form is:
 \beqn \label{eq:equivalencedef1}
 & \ & 
 \!\!\!\!\!\!\!\!\!\!\!\!\!\!\!\!
  \forall \ 0 \le a\le b:
   \nonumber \\ & \ & 
\!\!\!\!\!\!\!\!\!\!\!\!\!\!\!\!
\!\!\!\!\!\!\!\!
E(x_b)- E(x_a) + \int_{a}^{b} \left( \frac{1}{2} \abs{x_t'}^2 +\frac{1}{2} \abs{\nabla E}^2(x_t) \right) dt  \le 0.
 \eeqn
 The advantage of formulation \eqref{eq:equivalencedef1} is that it only uses
 quantities that can be defined in a metric space (see below for definition of $\abs{x_t'}$ and $ \abs{\nabla E}$).
The corresponding computation for a bi-variate functional $\costmfg$ is:
 \beqn \label{eq:equivalencedef2}
 & \ & 
  \forall  \ 0 \le a\le b: \ \int_{a}^{b} \left( \frac{d}{dt} \costmfgx(x_t,\nu)\Big|_{\nu=y_t} \right) dt
   \nonumber \\ & \ & 
   +  \int_{a}^{b}  \left(\frac{1}{2} \abs{x_t'}^2 +\frac{1}{2} \abs{\nabla_1 \costmfgx}^2(x_t,y_t) \right) dt 
 \nonumber \\ & \ & 
+ \int_{a}^{b} \left( \frac{d}{dt} \costmfgy(\nu,y_t)\Big|_{\nu=x_t} \right) dt 
 \nonumber \\ & \ & 
+  \int_{a}^{b}  \left(\frac{1}{2} \abs{y_t'}^2 +\frac{1}{2} \abs{\nabla_2 \costmfgy}^2(x_t,y_t) \right) dt \le 0.
 \eeqn
However this formulation poses specific problems 
as in general the solution $(x_t,y_t)_{t\ge 0}$ is only absolutely continuous (with respect to time) while, for instance, the manipulation of 
the term $\frac{d}{dt} \costmfgx(x_t,\nu)\Big|_{\nu=y_t} $ requires additional assumptions. 
This will be made precise later.


\subsection{Definition of (EDI style) equilibrium flows}

Let us recall the following definition:

\begin{definition}
	A curve $x:[0,T]\to (\espacex,d)$ is called absolutely continuous if there exists $f\in L^1(0,T)$ such that 
	\beq \label{eq:defabsont}
	d(x_{t_1},x_{t_2})\le \int_{t_1}^{t_2} f(t) dt, \ \forall {t_1} < {t_2}, \ {t_1},{t_2} \in [0,T].
	\eeq 
\end{definition}
For an absolutely continuous curve $(x_t)_{t\in [0,T]}$ the metric derivative of $x$ at $r$ defined by 
\beq
\abs{x_r'} = \lim_{h \to 0} \frac{d(x_{r+h},x_r)}{\abs{h}},
\eeq
exists a.e.,  belongs to  $L^1(0,T)$ and is the smallest $L^1$ function that verifies \eqref{eq:defabsont}.

We suppose from now on that $\costmfg$ satisfies the assumption:
\begin{enumerate}[label={\bf(A$_{\arabic*}$)}]	
	\item There exists $C_1 < \infty$ such that 
	$\costmfgfx{y}{x}, \costmfgfy{y}{x} \ge -C_1$,	
	$\forall x,y \in \espacex \times \espacey$. 
	\label{item:hypborneinf}
\end{enumerate}

For any $\alpha, \beta \in \R $, $\alpha \le \beta$, we denote by $\division{\alpha}{\beta}$ the set of divisions of the interval $[\alpha,\beta]$. 
Let $z=(x,y)=(x_t,y_t)_{t \in [0,T]} $ be an absolutely continuous curve in $\espacex \times \espacey$; define for $0 \le a \le b \le T$ and a division $\Delta=\{ a=t_0< t_1 < ... t_{N_\Delta} = b \} \in \division{a}{b}$:
\beq
\varxx(\Delta;z,a,b)= \sum_k \costmfgfx{x_{t_{k+1}}}{y_{t_{k}}}- \costmfgfx{x_{t_{k}}}{y_{t_{k}}}.
\eeq
\beq \label{eq:defvarx2}
\varxx(z,a,b) = \liminf_{\Delta \in \division{a}{b}, \ |\Delta| \to 0}  \varxx(\Delta;z,a,b).
\eeq

Similar definitions are introduced for $\varxy(z,a,b)$ (summing the variations of $\costmfgy$ along the curve $z$). Furthermore we denote
\beq
\varx(\Delta;z,a,b)= \varxx(\Delta;z,a,b) +\varxy(\Delta;z,a,b).
\eeq

\begin{remark} When $\espacex$ is e.g., Euclidian and under regularity assumptions on $\costmfg$ 
	it is easy to check that   
	$\varxx(x,a,b) = \int_{a}^{b} \frac{d}{dt} \costmfgx(x_t,\nu)\Big|_{\nu=y_t} dt$ and the same for $\varxy$.
\end{remark}

We are now ready to state the formal definition of a solution of \eqref{eq:eqflow0} in the abstract setting of metric spaces. The particular flavor we use
is the so-called "EDI" solution, see \cite{ambrosio_gradient_2008} for details.

\begin{definition}[EDI equilibrium flow] \label{def:edieqflow}
An absolutely continuous curve $z=(x_t,y_t)_{t\in[0,T]}$ is called an EDI-equlibrium flow 
starting from $(\barx,\bary)$ if $\lim_{t \to 0} (x_t,y_t) = (\barx,\bary)$ and:
\beqn
\label{eq:defEDIsolution1}
& \ & 
\displaystyle \forall s\geq 0,\ 
\varx(z,0,s)
+\frac{1}{2}\int_0^s \abs{x_r'}^2 + \abs{y_r'}^2\,\mathrm{d}r
\nonumber \\ & \ & 
+\frac{1}{2}\int_0^s\abs{ \nabla_1 \costmfgx }^2(x_r,y_r) + \abs{ \nabla_2 \costmfgy }^2(x_r,y_r)\,\mathrm{d}r\leq 0,
\\ 
& \ & 
\textrm{a.e.}\ t>0,\ \forall s\geq t,\ \varx(x,t,s)+\frac{1}{2}\int_t^s\abs{x_r'}^2 + \abs{y_r'}^2\,\mathrm{d}r
\nonumber \\ & \ & 
+\frac{1}{2}\int_t^s\abs{\nabla_1 \costmfg}^2(x_r,x_r)+ \abs{ \nabla_2 \costmfgy }^2(x_r,y_r)\,\mathrm{d}r\leq 0,
\nonumber
\\ & \ & \ 
\label{eq:defEDIsolution2}
\eeqn
where 
 the slope $\abs{\nabla_1 \costmfgx}(x,y)$ of $\costmfgx(\cdot,\cdot)$  with respect to the first argument evaluated at $(x,y)$ is:
\beq
\abs{\nabla_1 \costmfgx}(x,y)=\underset{u\to x}{\limsup}\,\frac{(\costmfgx(x,y)-\costmfgx(u,y))^+}{d(x,u)},\eeq
and similarly for $\abs{\nabla_2 \costmfgy}(x,y)$.
\end{definition}

\begin{remark}
For the particular case of a Hilbert space the definition above coincides with the usual definition of an evolution equation \eqref{eq:eqflow0}.
\end{remark}

\subsection{Convergence of numerical schemes}

	Let us denote 
\beqn \label{eq:minimizer}
& \ & 
	\minimizer^x(x,y,\tau)=
	arg min_{u \in \espacex} \frac{d(u,x)^2}{2 \tau} + \costmfgfx{u}{y}
	\\ & \ & 
	\minimizer^y(x,y,\tau)=
	arg min_{u \in \espacey} \frac{d(u,y)^2}{2 \tau} + \costmfgfy{x}{u}.	
\eeqn
	With this definition the  numerical scheme in equations \eqref{eq:bestreplydef}-\eqref{eq:bestreplydefy}
can be written as 
\beq
x_{k+1}^\tau \in \minimizer^x(x_{k}^\tau,y_{k}^\tau,\tau), 
y_{k+1}^\tau \in \minimizer^y(x_{k+1}^\tau,y_{k}^\tau,\tau).
\eeq

The goal of this contribution is to investigate  whether when $\tau \to 0$ the set $\{ (x_k^\tau,y_k^\tau), k\ge 1 \}$ converges to a solution of 
\eqref{eq:eqflow0} as defined in  \eqref{eq:defEDIsolution1}-\eqref{eq:defEDIsolution2}.

In order to work with meaningful objects, we introduce the following assumption which is the analogue of \cite[Assumption 4.8 page 67]{Ambrosio2013}:

\begin{enumerate}[resume,label={\bf(A$_{\arabic*}$)}]	
\item	There exists $\bartau> 0$ such that for any $\tau \le \bartau$ and $(x,y) \in \espacex \times \espacey$: 
\beq \label{eq:existencediscrete}
 \minimizer^x(x,y,\tau) \neq \emptyset, 
 \minimizer^y(x,y,\tau) \neq \emptyset.
\eeq
	\label{item:hypexistencediscrete}
\end{enumerate}

Assuming that assumption \ref{item:hypexistencediscrete} is satisfied, we can define the interpolation {\it \`a la} de Giorgi
 which is a curve 
$t \in [0,T] \mapsto (x^\tau_t,y^\tau_t)$ such that $(x^\tau_0,y^\tau_0)= (\barx,\bary)$ and
$ \forall t \in ]k \tau, (k+1)\tau ]$:
\beq
x^\tau_t \in \minimizer^x(x^\tau_{k \tau},y^\tau_{k \tau},t-k\tau),
y^\tau_t \in \minimizer^y(x^\tau_{(k+1) \tau},y^\tau_{k \tau},t-k\tau).
\label{eq:definterp} 
\eeq

We will need some additional hypothesis:

\begin{enumerate}[resume,label={\bf(A$_{\arabic*}$)}]	
\item For any $c \in \R$, $r >0$ and $(x,y)\in \espacex \times \espacey$ the sets
$\{ u \in \espacex | \costmfgfx{u}{y} \le c, d(u,x)\le r \}$ 
and
$\{ u \in \espacey | \costmfgfy{x}{u} \le c, d(u,u)\le r \}$ 
are both compact.
\label{item:hypcompactlevels}
\item The slopes $\abs{\nabla_1 \costmfgx}$ and 
$\abs{\nabla_2 \costmfgy}$
are lower semicontinuous. 
\label{item:hypslopelsc}
\item The function
$\costmfgx$ is
Lipschitz  with respect to the second argument
and  $\costmfgy$  is
Lipschitz  with respect to the first argument. 
\label{item:hyplipschitzsecondargument}
%
\end{enumerate}

\begin{enumerate}[resume,label={\bf(A$_{\arabic*}$)}]	
	\item For any absolutely continuous curve $z=(x_t,y_t)_{t\in [a,b]}$:
	\label{item:hyplimitupsilon}
\beq \label{eq:hyp}
\!\!\!\!\!\!\!\!\!\!\!\!\!\!\!  
\varx(x,a,b) \le \liminf_{
	\begin{matrix}
	|\Delta_n| \to 0 \\ z_n=(x_n,y_n) \to z \\
\sup_n \int_{a}^{b} \abs{ \dot{x}_n(t) } + \abs{ \dot{y}_n(t) } dt < \infty 
\end{matrix}	
}  
\varx(\Delta_n;z_n,a,b), 
\eeq
	where the convergence of the curves $z_n$ to $z$ is in the uniform (on compacts) norm.
\end{enumerate}

\begin{enumerate}[resume,label={\bf(A$_{\arabic*}$)}]	
\item \label{item:hypdouble}
There exists $C_L< \infty$ such that for any $x,y,u,w$:
\beq \label{eq:hypdouble}
\!\!\!\!\!\!\!\!\!\!\!\!
\abs{\costmfgfx{u}{v} + \costmfgfx{u}{y}- \costmfgfx{w}{v}-\costmfgfx{w}{y} }\le C_L d(u,w)d(v,y).
\eeq
and the same for $\costmfgfy{u}{v}$.
\end{enumerate}

\begin{remark} \label{rem:a7a8}
		The assumption \ref{item:hypdouble}  implies  \ref{item:hyplimitupsilon}
(see \cite[Lemma 2]{turinici_metric_2017}).
\end{remark}

With these provisions, the properties of the curves obtained by the numerical scheme \eqref{eq:bestreplydef} are detailed in the Theorem \ref{thm:edi}.

\begin{theorem} \label{thm:edi}
	Let $\costmfg$ satisfying assumptions  \ref{item:hypborneinf}, 	\ref{item:hypexistencediscrete}, \ref{item:hypcompactlevels}, \ref{item:hypslopelsc}, 
	\ref{item:hyplipschitzsecondargument} and \ref{item:hyplimitupsilon}.
Then	the set of curves $\{ (x_t^\tau,y_t^\tau)_{t \in [0,T]}; \tau \le \bar{\tau} \}$ defined in ~\eqref{eq:definterp} 
	 is relatively compact in the set of curves in 
	 $\espacex \times \espacey$ with local uniform convergence and any limit curve
	is an EDI equilibrium flow in the sense of Definition~\ref{def:edieqflow}.	
\end{theorem} 

\bproof 
The proof is somehow technical but is a adaptation of the proof of Theorem 1 in \cite{turinici_metric_2017}: first we show that the map
$\tau \mapsto \frac{d(x_\tau,x)^2}{2\tau} + \costmfgfx{x_\tau}{y} $ is locally Lipshitz. Then the discrete identity is obtained as in 
\cite[formula (32)]{turinici_metric_2017} and then estimations similar to \cite[formulas (35) and (36)]{turinici_metric_2017} allow to conclude.
\eproof

Similar results hold for the convex case (see \cite[Theorem 2]{turinici_metric_2017}).

\section{Applications}
Using Lemma  \ref{lemma:explicit} we conclude that if in some circumstances there are 
ways to use explicit numerical schemes (like for GANs), the convergence is also ensured for the explicit schemes, once the implicit ones converge.

Let us now inquire what are the consequences of the theoretical results for WGAN training. 
Because the Discriminator is trained first one can consider the variable $x$ to be the Discriminator network parameters 
that will result in a Lipschitz function $\ell= \ell(x)$ and $y$ to be the Generator parameters that will generate a distribution $\mu= \mu(y)$.
Functions 
$\costmfgx$ and $\costmfgy$ are given by \eqref{eq:ganfunctions} thus in particular 
$\costmfgx=- \costmfgy$. With these notations we can apply the Theorem \ref{thm:edi} and obtain that, in the limit of a vanishing learning rate, the WGAN training 
will tend to some evolution curve, both in the space $\espacex \times \espacey$ of parameters, but also in the space of the 
distributions (where  $\mu$ belongs) and Lipschitz function (where $\ell$ belongs).

As a further application, we can investigate the conditions under which the GAN training give rise to a mode collapse. 
The discussion below is not a mathematical proof but oriented towards a practical understanding. 
A mode collapse describes, e.g., the situation when
a strong Discriminator pushes the Generator to only produce a limited number of samples with a loss in diversity.
With our notations, this means that the evolution $(x_t, y_t)$ will be close to a constant $(x^\infty,y^\infty)$ 
but the corresponding distribution $\mu(y^\infty)$ is far from $\mu_r$ but can be expressed (at least approximately) as a finite sum of Dirac masses
$\sum_{a=1}^{A} p_a \delta_{i_a}$ where $i_a \in \Omega$ are  given images. Using the insight from Lemma \ref{lem:objectevolution}
and under assumption that the generator network is locally injective (i.e., does not generate redundant probability laws) the point  $\mu(y^\infty)$ is a critical point of the 
loss function. But, denoting $l^\infty$ the Lipschitz function corresponding to the Discriminator network, the loss function for the Generator will be 
$\E_{\mu}(\ell^\infty)-\E_{\mu_r}(\ell^\infty)$ and the loss of the Discriminator will be 
$\ell \mapsto \E_{\mu_r}(\ell)-\E_{\mu^\infty}(\ell)$. 
For such a loss function, the information that $\mu$ is a sum of Dirac masses and also a critical point of 
$\mu \mapsto \E_{\mu}(\ell^\infty)-\E_{\mu_r}(\ell^\infty)$
implies that moving (in the space of probability laws endowed with the $1$-Wasserstein metric) towards any other Dirac mass does not change (decrease) the loss value (to the first order). Therefore the 
images $i_a$ in the support of the measure $\mu$ are necessarily of lowest possible loss value i.e., if the discriminator is good enough, are members of the original "real" image values (in mathematical terms are members of the support of $\mu_r$). On the other hand, if the dynamics of the Discriminator  is also blocked in some point $\ell^\infty$, this means again that, to the first order,
$\E_{\mu^\infty}(\ell)-\E_{\mu_r}(\ell)$ cannot be increased locally when $\ell$ is slightly perturbed around $\ell^\infty$. Or, since $\mu$ and $\mu_r$ are different we obtain a contradiction. Therefore the mode collapse is {\bf not}
a legitimate limit dynamics. 
We can therefore conclude that if mode collapse happens this is due to a too large time step, to a not strong enough Generator architecture or to numerical traps that can be removed by perturbating slightly the dynamics.

\section{Discussion and conclusion}

When averaging out the steps of a SGD one obtains the gradient flow of the loss functional. The question that we ask in this paper is what is obtained when one averages out the generator-discriminator dynamics encountered in GANs. To answer the question we notice that in GANs the ground metric is not always of $L^2$ type but can be arbitrary (Wasserstein metric as in \cite{wasswassgan}, Banach norm as in \cite{adler_banach_2018}, etc). Thus we re-formulate the question: when the learning rate becomes smaller and smaller, is there any limit for the curves obtained during the GAN training ? Does this correspond to a dynamical system ? 
We first give sufficient conditions for this convergence in general metric space when the learning process is composed of implicit steps. On the other side we recall that under mild conditions explicit and implicit steps will be arbitrary close thus converge to the same limit. Therefore the dynamics of GAN training will in general follow the solution of a evolution equation whose details are given explicitly in equation \eqref{eq:eqflow0}. The knowledge of such a fact can help better understand the GAN optimization dynamics and the mode collapse phenomena.

\section*{Acknowledgements}

\appendix
\section{Appendix}
\subsection{Explicit and implicit numerical schemes in Hilbert spaces}

We recall below a standard result on the relationship between explicit 
and implicit numerical schemes in Hilbert spaces.
\begin{lemma} 
	Let $H$ be a Hilbert space, $f:H \to \R$ a bounded Lipschitz function with Lipschitz constant $L$ 
	and two numerical schemes defined by the recurrences:
	\begin{eqnarray}
	& \ & 
	x_{n+1}^E = x_{n}^E + \tau f(x_n^E), \ \ x_0^E = \barx 
	\\
	& \ & 
	x_{n+1}^I = x_{n}^I + \tau f(x_{n+1}^I), \ \ x_0^I = \barx
	\label{eq:implicitgen}
	\end{eqnarray}
	Then for $\tau$ small enough:
	\begin{enumerate}
		\item \label{item:lemmaimplicit1} the implicit scheme \eqref{eq:implicitgen} has a unique solution
		for any step $n \ge 0$.
		\item \label{item:lemmaimplicit2} let $T= N \tau$ for some fixed $N$, then 
		$\| x_{N}^I - x_{N}^E \| \le  C \tau $, with the constant $C$ depending only on $f$, $\barx$
		and $T$.
	\end{enumerate}
	\label{lemma:explicit}
\end{lemma}
\begin{remark}	
	Note that point \ref{item:lemmaimplicit2} implies in particular that if, for $\tau \to 0$, the implicit curves 
	$(x_n^I)_{n\ge 0}$
	converge to some limit curve then the explicit curves 
	$(x_n^E)_{n\ge 0}$ converge to the same. However in order to avoid technicalities we will not state precisely what the full curves are and what kind of convergence is obtained.
\end{remark}

\bproof Point \ref{item:lemmaimplicit1} is obtained by a Picard procedure after observing that the mapping
$x \mapsto x_{n}^E + \tau f(x)$ is a contraction for $\tau$ small enough.
For the point \ref{item:lemmaimplicit2} we make use of the Lipschitz constant of $f$:
\begin{eqnarray}
& \ & 
\| x_{n+1}^I - x_{n+1}^E \| \le \| x_{n}^I - x_{n}^E \| + \tau L  \| x_{n+1}^I - x_{n}^E \|
\nonumber \\
& \ & 
\le \| x_{n}^I - x_{n}^E \| + \tau L  \left(\| x_{n}^I - x_{n}^E \| + \tau f( x_{n+1}^I ) \right)
\end{eqnarray}
Thus, denoting by $C_f$ an upper bound on $f$: 
\begin{equation}
\| x_{n+1}^I - x_{n+1}^E \| 
\le (1+\tau L) \| x_{n}^I - x_{n}^E \| + \tau^2 L  C_f. 
\end{equation}

If suffices now to use the discrete version of the Gronwall lemma to obtain
$\| x_{n+1}^I - x_{n+1}^E \| \le e^{\tau L} (n+1) \tau^2 L C_f$, and the conclusion follows from:
\begin{equation}
\| x_{N}^I - x_{N}^E \| \le \tau e^{\tau L} T  L C_f.
\end{equation}

\subsection{Critical points of gradient flows on intermediary spaces}

We investigate in this section a simple situation of a gradient flow of a composed function. Suppose thus 
an initial space $X_p=\R^n$ (for a GAN the neural network parameter space), an object space $X_o=\R^m$ (for a GAN this will be the space of probability measures 
where $\mu$ belongs and  that of Lipschitz functions where $\ell$ belongs). Consider also two functions 
$g:X_p \to X_o$, $f : X_o \to \R$ and the gradient flow:
\begin{equation}
x_t' = - \nabla_x (f \circ g)(x_t),
\label{eq:composedflow}
\end{equation}
where for any function we denote by $\nabla$ its differential; for instance $\nabla_o f(o)$ is the gradient of $f$ at the point $o$, taken as a row vector, 
$\nabla_x g(x)$ is the $m \times n$ Jacobian matrix of $g$ at $x$ (entry $i,j$ being $\partial g_i / \partial x_j$).

This dynamics in parameter space $X_p$ defines a dynamics $o_t = g(x_t)$ in the object space $X_o$.
We want to investigate the relationship between the dynamics $x_t$ and $o_t$ when the evolution 
\eqref{eq:composedflow} ends up in a stationary point i.e., stalls at some given point $x^\infty \in X_p$ and 
$o^\infty= g(x^\infty) \in X_o$.

\begin{lemma} 
\noindent Suppose that the functions $f$ and $g$ are of $C^1$ class (i.e. with continuous derivatives). Then: 
\begin{enumerate}
	\item denoting $o_t = g(x_t)$ the dynamics in parameter space can also be written
\begin{equation}
x_t' = -  (\nabla_o f)(o_t)  \cdot (\nabla_x g)(x_t).
\label{eq:composedflow2}
\end{equation}	
\item the dynamic in object space $X_o$ is:
\begin{equation}
o_t' = -  (\nabla_o f)(o_t) \cdot  (\nabla_x g)(x_t)  \cdot (\nabla_x g)(x_t)^T .
\label{eq:objectflow}
\end{equation}
In particular the dynamics in object space is not in general a gradient flow (but will be when $(\nabla_x g) \cdot (\nabla_x g)^T = Id$).
\newcounter{enumTemp} \setcounter{enumTemp}{\theenumi} \end{enumerate}

\noindent Suppose now that the dynamics \eqref{eq:composedflow} is such that for some $t\ge t_1$ we have 
	$x_t = x^\infty$. Then:
\begin{enumerate} \setcounter{enumi}{\theenumTemp}
\item $ o^\infty=g(x^\infty)$ is a critical point of $f$ (i.e.,  $\nabla_o f(o^\infty)=0$) as soon as $\nabla_o g ^T$ is locally injective around $o^\infty$ 
(which implies that $g$ is locally injective around $o^\infty$).
\end{enumerate}
\label{lem:objectevolution}	
\end{lemma} 
\bproof 
Formulas  \eqref{eq:composedflow2}  and \eqref{eq:objectflow} are derived from  \eqref{eq:composedflow} using the chain rule. 
Suppose now $x^\infty$ is a critical stationary point of the dynamics  \eqref{eq:composedflow}. This of course implies that $o_t= g(x_t)$ will also
be constant but is not enough to conclude that $o^\infty$ is a critical point of $f$. But, since the time derivative $x_t'$ in  the evolution equation is zero 
for the constant dynamics $x_t = x^\infty$, 
we have that 
$ (\nabla_o f)(o^\infty) \cdot (\nabla_x g)(x^\infty)= 0$, which, when $(\nabla_x g)^T(x^\infty)$ is injective will imply that
$(\nabla_o f)(o^\infty)= 0$.
\eproof

Note that by the rank theorem the local injectivity of $(\nabla_x g)^T(x^\infty)$ and of $g$ are related.


\begin{thebibliography}{10}
	\providecommand{\url}[1]{\texttt{#1}}
	\providecommand{\urlprefix}{URL }
	\providecommand{\doi}[1]{https://doi.org/#1}
	
	\bibitem{adler_banach_2018}
	Adler, J., Lunz, S.: Banach {Wasserstein} {GAN}. In: Bengio, S., Wallach, H.,
	Larochelle, H., Grauman, K., Cesa-Bianchi, N., Garnett, R. (eds.) Advances in
	{Neural} {Information} {Processing} {Systems} 31, pp. 6754--6763. Curran
	Associates, Inc. (2018),
	\url{http://papers.nips.cc/paper/7909-banach-wasserstein-gan.pdf}
	
	\bibitem{Ambrosio2013}
	Ambrosio, L., Gigli, N.: Modelling and Optimisation of Flows on Networks:
	Cetraro, Italy 2009, Editors: Benedetto Piccoli, Michel Rascle, chap. A
	User's Guide to Optimal Transport, pp. 1--155. Springer Berlin Heidelberg,
	Berlin, Heidelberg (2013)
	
	\bibitem{ambrosio_gradient_2008}
	Ambrosio, L., Gigli, N., Savaré, G.: Gradient flows in metric spaces and in
	the space of probability measures. 2nd ed. Basel: Birkhäuser, 2nd ed. edn.
	(2008)
	
	\bibitem{gigli}
	{Ambrosio}, L., {Gigli}, N., {Savar\'e}, G.: {Gradient flows in metric spaces
		and in the space of probability measures. 2nd ed.} Basel: Birkh\"auser, 2nd
	ed. edn. (2008)
	
	\bibitem{arjovsky_wasserstein_2017}
	Arjovsky, M., Chintala, S., Bottou, L.: Wasserstein gan. arXiv preprint
	arXiv:1701.07875  (2017)
	
	\bibitem{Blanchet2014}
	Blanchet, A., Carlier, G.: Remarks on existence and uniqueness of cournot--nash
	equilibria in the non-potential case. Mathematics and Financial Economics
	\textbf{8}(4),  417--433 (2014). \doi{10.1007/s11579-014-0127-z},
	\url{http://dx.doi.org/10.1007/s11579-014-0127-z}
	
	\bibitem{brock_large_2019}
	Brock, A., Donahue, J., Simonyan, K.: Large {Scale} {GAN} {Training} for {High}
	{Fidelity} {Natural} {Image} {Synthesis}. arXiv:1809.11096 [cs, stat]  (Feb
	2019), \url{http://arxiv.org/abs/1809.11096}, arXiv: 1809.11096
	
	\bibitem{metricbook}
	Burago, D., Burago, Y., Ivanov, S.: A course in metric geometry, Graduate
	Studies in Mathematics, vol.~33. American Mathematical Society, Providence,
	RI (2001). \doi{10.1090/gsm/033}, \url{http://dx.doi.org/10.1090/gsm/033}
	
	\bibitem{deshpande_max-sliced_2019}
	Deshpande, I., Hu, Y.T., Sun, R., Pyrros, A., Siddiqui, N., Koyejo, S., Zhao,
	Z., Forsyth, D., Schwing, A.: Max-{Sliced} {Wasserstein} {Distance} and its
	use for {GANs}. arXiv preprint arXiv:1904.05877  (2019)
	
	\bibitem{wasswassgan}
	Dukler, Y., Li, W., Lin, A.T., MontÃºfar, G.: {Wasserstein of Wasserstein
		loss for learning generative models}. In: Chaudhuri, K. (ed.) Proceedings of
	the 36th international conference on machine learning, 9-15 June 2019, Long
	Beach, California, USA, Proceedings of machine learning research, vol.~97,
	pp. 1716--1725. PMLR, Long Beach, California (2019)
	
	\bibitem{Ferreira}
	Ferreira, L.C.F., Valencia-Guevara, J.C.: Gradient flows of time-dependent
	functionals in metric spaces and applications to pdes. Monatshefte f{\"u}r
	Mathematik pp. 1--38 (2017). \doi{10.1007/s00605-017-1037-y},
	\url{http://dx.doi.org/10.1007/s00605-017-1037-y}
	
	\bibitem{gulrajani_improved_2017}
	Gulrajani, I., Ahmed, F., Arjovsky, M., Dumoulin, V., Courville, A.C.: Improved
	training of {Wasserstein GANs}. In: Advances in neural information processing
	systems. pp. 5767--5777 (2017)
	
	\bibitem{pix2pixGAN}
	Isola, P., Zhu, J.Y., Zhou, T., Efros, A.A.: Image-to-{Image} {Translation}
	with {Conditional} {Adversarial} {Networks}. arXiv:1611.07004 [cs]  (Nov
	2018), \url{http://arxiv.org/abs/1611.07004}, arXiv: 1611.07004
	
	\bibitem{iwaki_implicit_2019}
	Iwaki, R., Asada, M.: Implicit incremental natural actor critic algorithm.
	Neural Networks  \textbf{109},  103--112 (Jan 2019).
	\doi{10.1016/j.neunet.2018.10.007},
	\url{http://www.sciencedirect.com/science/article/pii/S0893608018302922}
	
	\bibitem{jko}
	Jordan, R., Kinderlehrer, D., Otto, F.: The variational formulation of the
	{F}okker-{P}lanck equation. SIAM J. Math. Anal.  \textbf{29}(1),  1--17
	(1998). \doi{10.1137/S0036141096303359},
	\url{http://dx.doi.org/10.1137/S0036141096303359}
	
	\bibitem{karras_progressive_2018}
	Karras, T., Aila, T., Laine, S., Lehtinen, J.: Progressive {Growing} of {GANs}
	for {Improved} {Quality}, {Stability}, and {Variation}. In: International
	{Conference} on {Learning} {Representations} (2018),
	\url{https://openreview.net/forum?id=Hk99zCeAb}
	
	\bibitem{stylegan}
	Karras, T., Laine, S., Aila, T.: A {Style}-{Based} {Generator} {Architecture}
	for {Generative} {Adversarial} {Networks}. arXiv:1812.04948 [cs, stat]  (Mar
	2019), \url{http://arxiv.org/abs/1812.04948}, arXiv: 1812.04948
	
	\bibitem{kodali_convergence_2017}
	Kodali, N., Abernethy, J., Hays, J., Kira, Z.: On convergence and stability of
	{GANs}. arXiv preprint arXiv:1705.07215  (2017)
	
	\bibitem{kolouri_sliced-wasserstein_2018}
	Kolouri, S., Pope, P.E., Martin, C.E., Rohde, G.K.: Sliced-{Wasserstein}
	autoencoder: an embarrassingly simple generative model. arXiv preprint
	arXiv:1804.01947  (2018)
	
	\bibitem{evakopfer1}
	{Kopfer}, E.: {Gradient flow for the Boltzmann entropy and Cheeger's energy on
		time-dependent metric measure spaces}. ArXiv e-prints  (November 2016)
	
	\bibitem{coupled_gan}
	Liu, M.Y., Tuzel, O.: Coupled {Generative} {Adversarial} {Networks}.
	arXiv:1606.07536 [cs]  (Sep 2016), \url{http://arxiv.org/abs/1606.07536},
	arXiv: 1606.07536
	
	\bibitem{mielke2012}
	Mielke, A., Rossi, R., Savar\'e, G.: Variational convergence of gradient flows
	and rate-independent evolutions in metric spaces. Milan J. Math.
	\textbf{80}(2),  381--410 (2012). \doi{10.1007/s00032-012-0190-y},
	\url{http://dx.doi.org/10.1007/s00032-012-0190-y}
	
	\bibitem{mielke2013}
	Mielke, A., Rossi, R., Savar\'e, G.: Nonsmooth analysis of doubly nonlinear
	evolution equations. Calc. Var. Partial Differential Equations
	\textbf{46}(1-2),  253--310 (2013). \doi{10.1007/s00526-011-0482-z},
	\url{http://dx.doi.org/10.1007/s00526-011-0482-z}
	
	\bibitem{radford_unsupervised_2016}
	Radford, A., Metz, L., Chintala, S.: Unsupervised {Representation} {Learning}
	with {Deep} {Convolutional} {Generative} {Adversarial} {Networks}.
	arXiv:1511.06434 [cs]  (Jan 2016), \url{http://arxiv.org/abs/1511.06434},
	arXiv: 1511.06434
	
	\bibitem{mielke2008}
	Rossi, R., Mielke, A., Savar\'e, G.: A metric approach to a class of doubly
	nonlinear evolution equations and applications. Ann. Sc. Norm. Super. Pisa
	Cl. Sci. (5)  \textbf{7}(1),  97--169 (2008)
	
	\bibitem{filippo}
	Santambrogio, F.: Optimal transport for applied mathematicians. Progress in
	Nonlinear Differential Equations and their Applications, 87,
	Birkh\"auser/Springer, Cham (2015). \doi{10.1007/978-3-319-20828-2},
	\url{http://dx.doi.org/10.1007/978-3-319-20828-2}, calculus of variations,
	PDEs, and modeling
	
	\bibitem{sriperumbudur_hilbert_2010}
	Sriperumbudur, B.K., Gretton, A., Fukumizu, K., Schölkopf, B., Lanckriet,
	G.R.G.: Hilbert {Space} {Embeddings} and {Metrics} on {Probability}
	{Measures}. Journal of Machine Learning Research  \textbf{11}(Apr),
	1517--1561 (2010), \url{http://www.jmlr.org/papers/v11/sriperumbudur10a.html}
	
	\bibitem{turinici_metric_2017}
	Turinici, G.: Metric gradient flows with state dependent functionals: {The}
	{Nash}-{MFG} equilibrium flows and their numerical schemes. Nonlinear
	Analysis  \textbf{165},  163--181 (Dec 2017). \doi{10.1016/j.na.2017.10.002},
	\url{http://www.sciencedirect.com/science/article/pii/S0362546X17302444}
	
	\bibitem{Villani2009}
	Villani, C.: {Optimal transport. Old and new}, Grundlehren der mathematischen
	Wissenschaften, vol.~338. Springer (2009). \doi{10.1007/978-3-540-71050-9}
	
	\bibitem{wu_sliced_2019}
	Wu, J., Huang, Z., Acharya, D., Li, W., Thoma, J., Paudel, D.P., Gool, L.V.:
	Sliced {Wasserstein} generative models. In: Proceedings of the {IEEE}
	conference on computer vision and pattern recognition. pp. 3713--3722 (2019)
	
	\bibitem{zhu_unpaired_2017}
	Zhu, J.Y., Park, T., Isola, P., Efros, A.A.: Unpaired {Image}-{To}-{Image}
	{Translation} {Using} {Cycle}-{Consistent} {Adversarial} {Networks}. In: 2017
	IEEE International Conference on Computer Vision (ICCV). pp. 2223--2232
	(2017)
	
\end{thebibliography}
\end{document}